\documentclass[
]{ceurart}

\begin{document}
\copyrightyear{2021}
\copyrightclause{Copyright for this paper by its authors.
  Use permitted under Creative Commons License Attribution 4.0
  International (CC BY 4.0).}

\conference{Forum for Information Retrieval Evaluation, December 15-18, 2023, India}

\title{Multi-Label Classification of COVID-Tweets Using Large Language Models}

\author[1]{Aniket Deroy}[%
orcid=0000-0000-0000-0000,
email=roydanik18@kgpian.iitkgp.ac.in,
url=https://,
]
\address[1]{ IIT Kharagpur, Khargapur, India}

\author[1]{Subhankar Maity}[%
orcid=0000-0000-0000-0000,
email=subhankar.ai@kgpian.iitkgp.ac.in,
url=https://,
]

\begin{abstract}
Vaccination is important to minimize the risk and spread of various diseases. In recent years, vaccination has been a key step in countering the COVID-19 pandemic. However, many people are skeptical about the use of vaccines for various reasons, including the politics involved, the potential side effects of vaccines, etc. 
The goal in this task is to build an effective multi-label classifier to label a social media post (particularly, a tweet) according to the specific concern(s) towards vaccines as expressed by the author of the post.
We tried three different models-(a) Supervised BERT-large-uncased, (b) Supervised HateXplain model, and (c) Zero-Shot GPT-3.5 Turbo model. The Supervised BERT-large-uncased model performed best in our case. We achieved a macro-F1 score of 0.66, a Jaccard similarity score of 0.66, and received the sixth rank among other submissions.
Code is available at-\url{https://github.com/anonmous1981/AISOME}

\end{abstract}


\maketitle

\section{Introduction}

Vaccination, as a cornerstone of public health, plays a key role in reducing the risk and spread of various diseases. Over the years, vaccines have proven to be one of the most effective tools in combating infectious diseases, contributing significantly to global efforts to control and eradicate deadly pathogens. In recent times, the importance of vaccination has been underscored by the emergence of the COVID-19 pandemic, where vaccines have emerged as our most potent weapon in curbing the devastating impact of the virus (~\url{https://www.who.int/emergencies/diseases/novel-coronavirus-2019/covid-19-vaccines}). Beyond pandemic control, widespread vaccination is indispensable to prevent a spectrum of diseases, including those affecting vulnerable populations such as children and annual recurring threats such as influenza. Despite the undeniable success stories and scientific consensus surrounding vaccinations, a growing segment of the population remains skeptical about their use. 


Despite the undeniable benefits of vaccines, a growing phenomenon of vaccine hesitancy has emerged in recent years. Vaccine hesitancy refers to the delay in accepting or refusing vaccines, despite the availability of vaccination services. It is not limited to a specific demographic or geographic region, but is observed in diverse populations around the world.
Vaccine hesitancy can manifest itself in various forms, from the outright refusal of vaccines to concerns about vaccine safety, efficacy or mistrust in the motives of health authorities and pharmaceutical companies (~\url{https://www.ncbi.nlm.nih.gov/pmc/articles/PMC9351420/}). This hesitancy can have significant consequences, including reduced vaccination coverage rates, increased vulnerability to outbreaks, and a resurgence of preventable diseases.
The reasons behind vaccine hesitancy are complex and multifaceted. They often intersect with broader social issues, including the spread of misinformation, the mistrust of institutions, and political polarization. To effectively address vaccine hesitancy, it is imperative to understand the underlying factors that drive it.

This track~\cite{poddar2023aisome} is based on the CAVES dataset~\cite{poddar2022caves}.
Here, the goal is to build an effective multi-label classifier to label a social media post (particularly a tweet) according to the specific concern(s) towards vaccines as expressed by the author of the post.
We tried various classifiers, including BERT-large-uncased, and HateXplain after training on the AISOME training dataset.
We use GPT-3.5 Turbo in a zero-shot mode with prompt.
The results clearly show that the BERT-large-uncased model trained on the AISOME training data has performed best among all the models. Our team ranked 6th in this task.

\section{Problem Definition}
The goal is to build an effective multi-label classifier to label a social media post (particularly a tweet) according to the specific concern(s) for vaccines as expressed by the author of the post.

We consider the following concerns towards vaccines as the labels for the classification task: 
\begin{itemize}

\item \textbf{Unnecessary}: The tweet implies that vaccines may be unnecessary or posits the existence of more effective alternative treatments. 

\item \textbf{Mandatory}: Opposed to compulsory vaccination — The tweet implies that vaccines should not be required by law. 

\item \textbf{Pharma}: Opposed to Big Pharma — The tweet conveys the idea that large pharmaceutical companies are primarily motivated by profit, or it expresses a general distrust of such companies due to their historical actions. 

\item \textbf{Conspiracy}: Darker Conspiracy Angle — The tweet hints at a more intricate conspiracy beyond profit motives, such as the idea that vaccines might be used for surveillance or that COVID is being portrayed as a hoax. 

\item \textbf{Political}: The Political Aspect of Vaccination — The tweet raises concerns about the possibility of governments or politicians advancing their own interests through the promotion of vaccines.

\item \textbf{Country}: Originating country — The tweet expresses opposition to a vaccine due to the nation in which it was created or produced.

\item \textbf{Rushed}: Untested / Rushed Process — The tweet raises worries about the vaccines undergoing insufficient testing or questions the accuracy of the published data. 

\item \textbf{Ingredients}: The tweet raises issues regarding the components found in vaccines (e.g., fetal cells, chemicals) or the technology employed (e.g., the claim that mRNA vaccines have the potential to modify your DNA).

\item \textbf{Side-effect}: Adverse effects and fatalities — The tweet voices concerns regarding the vaccine's side effects, which include reported cases of deaths.

\item \textbf{Ineffective}: The tweet conveys concerns that the vaccines are not sufficiently effective and serve no practical purpose.

\item \textbf{Religious}: Religious grounds - Twitter opposes vaccines based on religious beliefs.

\item \textbf{None}: The tweet does not provide a particular explanation or offers an explanation different from the ones provided.

\end{itemize}
\section{Related Work}
\textbf{BERT} (Bidirectional Encoder Representations from Transformers)~\cite{devlin2018BERT} is a state-of-the-art natural language processing (NLP) model developed by Google in 2018. It has revolutionized the field of NLP with its innovative architecture and pre-training techniques. We used the \textbf{BERT-large-uncased} model for multilabel text classification in this work.

The \textbf{HateXplain} model~\cite{mathew2021hatexplain} works to address the complex issue of hate speech on social media platforms. The work seems to focus on multiple aspects of hate speech, including bias and interpretability, which are critical components in developing effective solutions to combat this problem. 

\textbf{GPT-3.5 Turbo}~\cite{NEURIPS2020_1457c0d6} is a highly advanced large language model (LLM) developed by OpenAI. It is part of the GPT-3 family of models and is known for its remarkable natural language understanding and generation capabilities. GPT-3.5 Turbo can comprehend and generate human-like text across a wide range of topics and tasks, making it a versatile tool for various applications, from chatbots and content generation to language translation, and more. This LLM is designed to assist with complex language-related tasks and provides impressive language generation capabilities based on a large amount of text data on which it was trained.
GPT-3.5 Turbo is used through prompting in a zero-shot mode.

This track is based on the work~\cite{poddar2022caves} which is a multi-label classification of COVID-related tweets with the aim of trying to improve the COVID vaccination process.

\section{Dataset}
We have received a training set of 9,921 tweets along with the corresponding labels.
We have received 486 tweets in the test set with labels.
We received a CSV file for training purposes that contained the ID, tweet, and label for 9,921 tweets.
We had received a CSV file for testing purposes that contained the ID, tweet, and label for 486 tweets.

\section{Methodology}
\subsection{Method 1}

\begin{figure}[h!]
  \centering
  \includegraphics[width=0.70\linewidth]{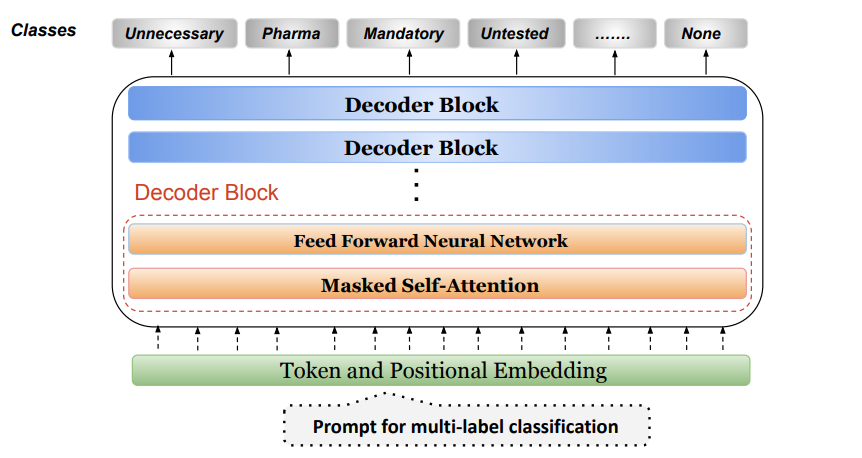}
  \caption{An overview of GPT for zero-shot multi-label classification.} \label{fig2}
\end{figure}

\begin{figure}[h!]
  \centering
  \includegraphics[width=\linewidth]{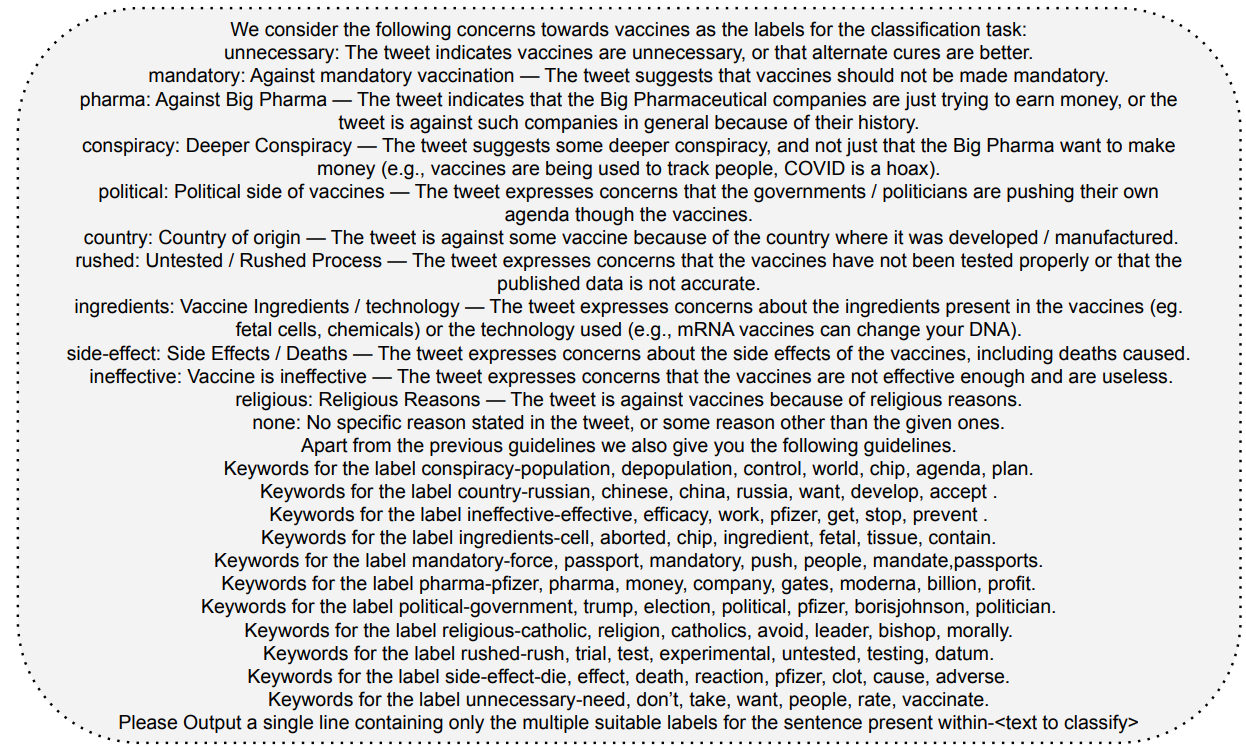}
  \caption{Prompt used for GPT-3.5 Turbo.} \label{fig3}
\end{figure}

We used the GPT-3.5 Turbo model in zero-shot mode. We give instructions to GPT-3.5 Turbo in zero-shot mode with the list of labels descriptions and the task to be performed. We also provide the list of the most important keywords corresponding to each label as instructions to the GPT-3.5 Turbo model. Then we provide every query to the model and ask it to provide corresponding labels for a multi-label classification problem. The hyperparameters are as follows: \texttt{temperature} = 0.7, \texttt{max-tokens} = 50, and \texttt{stop} = None.
A diagrammatic representation of the model is given in Figure~\ref{fig2}.
The prompt we use for the model is provided in Figure ~\ref{fig3}.

\subsection{Method 2} 
\begin{figure}[h!]
  \centering
  \includegraphics[width=0.70\linewidth]{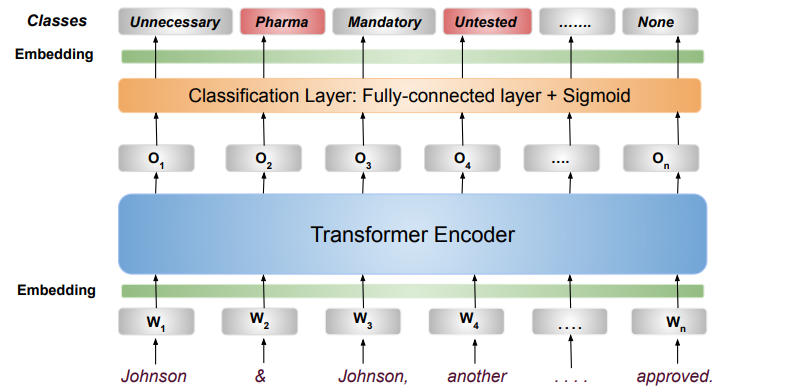}
  \caption{An overview of BERT for multi-label classification.} \label{fig1}
\end{figure}

We trained the BERT-large-uncased model using the training set of 9,921 tweets and the corresponding labels. We tested the model on the 486 tweets present in the test set. The embeddings from the BERT-large-uncased model were obtained and the embeddings were passed through a dense layer to obtain the final predictions. The dimensionality of the embeddings of the model is 1,024 and the maximum input token length is 512. The model is run for 100 epochs with a batch size of 1, a threshold of 0.5, and a learning rate of 2e-5. We use the sigmoid activation function. A diagrammatic representation of the model is shown in Figure~\ref{fig1}.
\\
\subsection{Method 3} We trained the HateXplain model using the training set of 9,921 tweets and corresponding labels. We tested the model on the 486 tweets present in the test set. The embeddings from the HateXplain model were obtained and the embeddings were passed through a dense layer to obtain the final predictions. The dimensionality of the embeddings of the model is 768 and the maximum input token length is 512. The model is run for 100 epochs at a batch size of 1, threshold of 0.5, and learning rate of 2e-5.
\\

\section{Results}
We observe that the BERT-large-uncased gives the best results when trained on the AISOME training dataset of
9,921 tweets along with their labels. 
Table~\ref{tab:t5} shows the results of all models in the AISOME test dataset for Macro-F1 and Jaccard Similarity. 

\begin{table}[h!]
    \caption{Result of all methods on the AISOME test dataset for Macro-F1 and Jaccard Similarity.}
     \label{tab:t5}
    \centering
     \renewcommand{\arraystretch}{1.13}   
\scalebox{0.80}{
    
    \begin{tabular}{ c|c|c|c|c|c }
    \toprule
         \textbf{Team\textunderscore{ID}} & \textbf{Summary of Methodology} & \textbf{Macro-F1} & \textbf{Jaccard Similarity} & \textbf{Rank} & \textbf{Run File}\\
         \midrule
         TextTitans & BERT-large-uncased (Method 2) & 0.66 & 0.66 & 6 & text\_titans\_social\_media2.csv\\ \midrule
         TextTitans & HateXplain (Method 3) & 0.54 & 0.57 & 21 & text\_titans\_hate\_explain2.csv\\ \midrule
        TextTitans & Zero Shot GPT-3.5 (Method 1) & 0.53 & 0.44 & 24 & text\_titans14.csv\\ \bottomrule

    \end{tabular}
  }  

\end{table}

\section{Conclusion and Future Work}
The task is focused on building a multi-label classifier that helps predict the nature of a tweet related to COVID or other disease-causing viruses.
We have tried various language models like the BERT-large-uncased, and HateXplain model after training on the AISOME training dataset and the GPT-3.5 model in a zero-shot setting by using prompt engineering.
The results show that the BERT-large-uncased (Method 2) has provided the best results. 
Future work will focus on increasing the amount of training data for training the models. In addition, a focus will be on trying several other large language models for the purpose of multilabel classification.


\bibliography{sample-ceur}

\begin{thebibliography}{5}
\expandafter\ifx\csname natexlab\endcsname\relax\def\natexlab#1{#1}\fi
\providecommand{\url}[1]{\texttt{#1}}
\providecommand{\href}[2]{#2}
\providecommand{\path}[1]{#1}
\providecommand{\DOIprefix}{doi:}
\providecommand{\ArXivprefix}{arXiv:}
\providecommand{\URLprefix}{URL: }
\providecommand{\Pubmedprefix}{pmid:}
\providecommand{\doi}[1]{\href{http://dx.doi.org/#1}{\path{#1}}}
\providecommand{\Pubmed}[1]{\href{pmid:#1}{\path{#1}}}
\providecommand{\bibinfo}[2]{#2}
\ifx\xfnm\relax \def\xfnm[#1]{\unskip,\space#1}\fi
\bibitem[{Poddar et~al.(2023)Poddar, Basu, Ghosh, and Ghosh}]{poddar2023aisome}
\bibinfo{author}{S.~Poddar}, \bibinfo{author}{M.~Basu},
  \bibinfo{author}{K.~Ghosh}, \bibinfo{author}{S.~Ghosh},
\newblock \bibinfo{title}{Overview of the fire 2023 track:artificial
  intelligence on social media (aisome)},
\newblock in: \bibinfo{booktitle}{Proceedings of the 15th Annual Meeting of the
  Forum for Information Retrieval Evaluation}, \bibinfo{year}{2023}.
\bibitem[{Poddar et~al.(2022)Poddar, Samad, Mukherjee, Ganguly, and
  Ghosh}]{poddar2022caves}
\bibinfo{author}{S.~Poddar}, \bibinfo{author}{A.~M. Samad},
  \bibinfo{author}{R.~Mukherjee}, \bibinfo{author}{N.~Ganguly},
  \bibinfo{author}{S.~Ghosh},
\newblock \bibinfo{title}{Caves: A dataset to facilitate explainable
  classification and summarization of concerns towards covid vaccines},
\newblock in: \bibinfo{booktitle}{Proceedings of the 45th International ACM
  SIGIR Conference on Research and Development in Information Retrieval},
  \bibinfo{year}{2022}, pp. \bibinfo{pages}{3154--3164}.
\bibitem[{Devlin et~al.(2018)Devlin, Chang, Lee, and
  Toutanova}]{devlin2018BERT}
\bibinfo{author}{J.~Devlin}, \bibinfo{author}{M.-W. Chang},
  \bibinfo{author}{K.~Lee}, \bibinfo{author}{K.~Toutanova},
\newblock \bibinfo{title}{Bert: Pre-training of deep bidirectional transformers
  for language understanding},
\newblock \bibinfo{journal}{arXiv preprint arXiv:1810.04805}
  (\bibinfo{year}{2018}).
\bibitem[{Mathew et~al.(2021)Mathew, Saha, Yimam, Biemann, Goyal, and
  Mukherjee}]{mathew2021hatexplain}
\bibinfo{author}{B.~Mathew}, \bibinfo{author}{P.~Saha}, \bibinfo{author}{S.~M.
  Yimam}, \bibinfo{author}{C.~Biemann}, \bibinfo{author}{P.~Goyal},
  \bibinfo{author}{A.~Mukherjee},
\newblock \bibinfo{title}{Hatexplain: A benchmark dataset for explainable hate
  speech detection},
\newblock in: \bibinfo{booktitle}{Proceedings of the AAAI conference on
  artificial intelligence}, volume~\bibinfo{volume}{35}, \bibinfo{year}{2021},
  pp. \bibinfo{pages}{14867--14875}.
\bibitem[{Brown et~al.(2020)Brown, Mann, Ryder, Subbiah, Kaplan, Dhariwal,
  Neelakantan, Shyam, Sastry, Askell, Agarwal, Herbert-Voss, Krueger, Henighan,
  Child, Ramesh, Ziegler, Wu, Winter, Hesse, Chen, Sigler, Litwin, Gray, Chess,
  Clark, Berner, McCandlish, Radford, Sutskever, and
  Amodei}]{NEURIPS2020_1457c0d6}
\bibinfo{author}{T.~Brown}, \bibinfo{author}{B.~Mann},
  \bibinfo{author}{N.~Ryder}, \bibinfo{author}{M.~Subbiah},
  \bibinfo{author}{J.~D. Kaplan}, \bibinfo{author}{P.~Dhariwal},
  \bibinfo{author}{A.~Neelakantan}, \bibinfo{author}{P.~Shyam},
  \bibinfo{author}{G.~Sastry}, \bibinfo{author}{A.~Askell},
  \bibinfo{author}{S.~Agarwal}, \bibinfo{author}{A.~Herbert-Voss},
  \bibinfo{author}{G.~Krueger}, \bibinfo{author}{T.~Henighan},
  \bibinfo{author}{R.~Child}, \bibinfo{author}{A.~Ramesh},
  \bibinfo{author}{D.~Ziegler}, \bibinfo{author}{J.~Wu},
  \bibinfo{author}{C.~Winter}, \bibinfo{author}{C.~Hesse},
  \bibinfo{author}{M.~Chen}, \bibinfo{author}{E.~Sigler},
  \bibinfo{author}{M.~Litwin}, \bibinfo{author}{S.~Gray},
  \bibinfo{author}{B.~Chess}, \bibinfo{author}{J.~Clark},
  \bibinfo{author}{C.~Berner}, \bibinfo{author}{S.~McCandlish},
  \bibinfo{author}{A.~Radford}, \bibinfo{author}{I.~Sutskever},
  \bibinfo{author}{D.~Amodei},
\newblock \bibinfo{title}{Language models are few-shot learners},
\newblock in: \bibinfo{editor}{H.~Larochelle}, \bibinfo{editor}{M.~Ranzato},
  \bibinfo{editor}{R.~Hadsell}, \bibinfo{editor}{M.~Balcan},
  \bibinfo{editor}{H.~Lin} (Eds.), \bibinfo{booktitle}{Advances in Neural
  Information Processing Systems}, volume~\bibinfo{volume}{33},
  \bibinfo{publisher}{Curran Associates, Inc.}, \bibinfo{year}{2020}, pp.
  \bibinfo{pages}{1877--1901}. \URLprefix
  \url{https://proceedings.neurips.cc/paper_files/paper/2020/file/1457c0d6bfcb4967418bfb8ac142f64a-Paper.pdf}.

\end{thebibliography}


\end{document}